\documentclass[11pt,english,draft,final,a4paper,twocolumn]{article}
\usepackage[ansinew]{inputenc}
\usepackage{babel}
\usepackage{graphicx}
\usepackage[small]{caption}

\usepackage{fixltx2e} % To prevent bad ordering of figure*

\usepackage{amsmath}
\usepackage{amsfonts}
\usepackage{amssymb}

\usepackage[usenames,dvipsnames]{color}
\usepackage[bookmarks=false]{hyperref}
\hypersetup{
colorlinks=true,       % false: boxed links; true: colored links
linkcolor=OliveGreen,          % color of internal links
citecolor=Sepia,        % color of links to bibliography
filecolor=magenta,      % color of file links
urlcolor=NavyBlue,          % color of external links
}
\usepackage[sort&compress]{natbib}
\usepackage{nicefrac}
\usepackage[cdot,squaren,derived]{SIunits}
\usepackage{subfig}

\newcommand{\ud}{\mathrm{d}}

\begin{document}
\title{\bf Modeling and frequency domain  analysis of nonlinear compliant joints for a passive dynamic swimmer}    % Supply information
\author{Juan Pablo Carbajal, Rafael Bayma, Marc Ziegler and Zi-Qiang Lang}              %   for the title page.

\date{\today}                           %   Use current date.
\maketitle                              % Print title page.
\setlength{\topmargin}{0cm}

\abstract{ In this paper we present the study of the mathematical model of a real life joint used in an underwater robotic fish. Fluid-structure interaction is utterly simplified and the motion of the joint is approximated by D\"uffing's equation. We compare the quality of analytical harmonic solutions previously reported, with the input-output relation obtained via truncated Volterra series expansion. Comparisons show a trade-off between accuracy and flexibility of the methods. The methods are discussed in detail in order to facilitate reproduction of our results. The approach presented herein can be used to verify results in nonlinear resonance applications and in the design of bio-inspired compliant robots that exploit passive properties of their dynamics. We focus on the potential use of this type of joint for energy extraction from environmental sources, in this case a K\'arm\'an vortex street shed by an obstacle in a flow. Open challenges and questions are mentioned throughout the document.}

\section{Introduction}
{\it How much of the diverse behavior we observe in animals is a direct expression of the dynamics of the individual's body?} In the last two decades, many characteristics of animal locomotion on land were successfully linked to the mechanical properties of the legs. The springy behavior observed in the trajectories of the center of mass during running, walking and jumping can be explained by the mechanical properties of the limbs and its tunning~(\cite{Farley93,Farley1996,Ferris1998,Dickinson2000,Kerdok2002,Moritz2003,Roberts2011}). The hypothesis that running, walking and jumping is tuned to the resonance of the underlying mechanical system is strongly supported by experimental evidence and by machines constructed based on this idea, the {\it passive dynamic walkers}~(\cite{Alexander1990,Ahlborn2002,McGeer1990,Thompson1989,Collins2005}). The consequence of such setting is energy efficient performance and alleviation of the controller.

Due to similarities between running and swimming~(\cite{Bejan06,Kokshenev2010}), it is not surprising (but not less exiting) that efficient locomotion in fluids was reported to relay on the dynamics of the body of the animal. Living trouts have been observed to exploit the energy in the flow they inhabit to reduce their swimming efforts~(\cite{Liao03}). Later, euthanized trouts performed passive self-propulsion when placed in the von K\'arm\'an vortex street shed by an obstacle in a flow~(\cite{Liao07}). The experimental results are supported by mathematical models, analytical and numerical~(\cite{Eldredge08,Kanso09,Alben09}). These models do not fully agree with each other, however this is expected due to the mathematical complexity of the interaction between structures and fluids, of which the passive case is the worst scenario: unprescribed motion of the interface boundary. This alone represents an open challenge for the mathematical modeling community. Lest the challenge remains unsolved for too long, robotic researchers design their machines with less fluid-dynamic rigor, pursuing the first {\it passive dynamic swimmer}. The final objective is to build a swimming machine that can perform at least as well as fish, and at comparable power ratings~(\cite{Harper97,Lauder07}).

For a robot to extract the energy in the surroundings, its mechanical properties have to be tuned to the environmental energy storage. Stated this way, the problem is one of energy harvesting, were nonlinear properties are believed to be beneficial~(\cite{Cottone09}). Therefore, we need to pin down the resonance characteristic of the actuators and joints to be used in the robot that, as their biological counterparts, are generally nonlinear. As if difficulty was lacking, the study of resonance of nonlinear systems has suffer a very slow developmental process that started early in the 1960's and has yet not overcome its infancy. In a technical report from 1958 by~\cite{Brilliant58} we read: \begin{quote}Sometimes nonlinearity is avoided, not because it would have an undesired effect in practice, but simply because its effect cannot be computed.\end{quote} Nowadays the situation is not completely different. However, we have more powerful computers, new simulation methods and some novel uses of classical tools promise to open the path ahead~\cite{Peng2007,Vakakis09}.

In this paper we present the study of the joint of a robot fish from a mathematical point of view. The amplitude of the oscillations of the joint in response to periodic forcing is studied. In section~\ref{sec:joint} we briefly introduce the real mechanical device and we move to its mathematical model in section~\ref{sec:model}. In the same section the model for fluid-structure interaction is briefly described. Approximated solutions methods are introduced in section~\ref{sec:SolMeth}. Results are presented in section~\ref{sec:orf}. Finally, we close the paper in section~\ref{sec:conclu} with a discussion on the implications of the results and the relevance of the approach to the design of robots and the test of controllers based on resonances.

\section{\label{sec:joint} A simple compliant joint}
\begin{figure*}[htb]
\centering
\includegraphics[width=.8\textwidth]{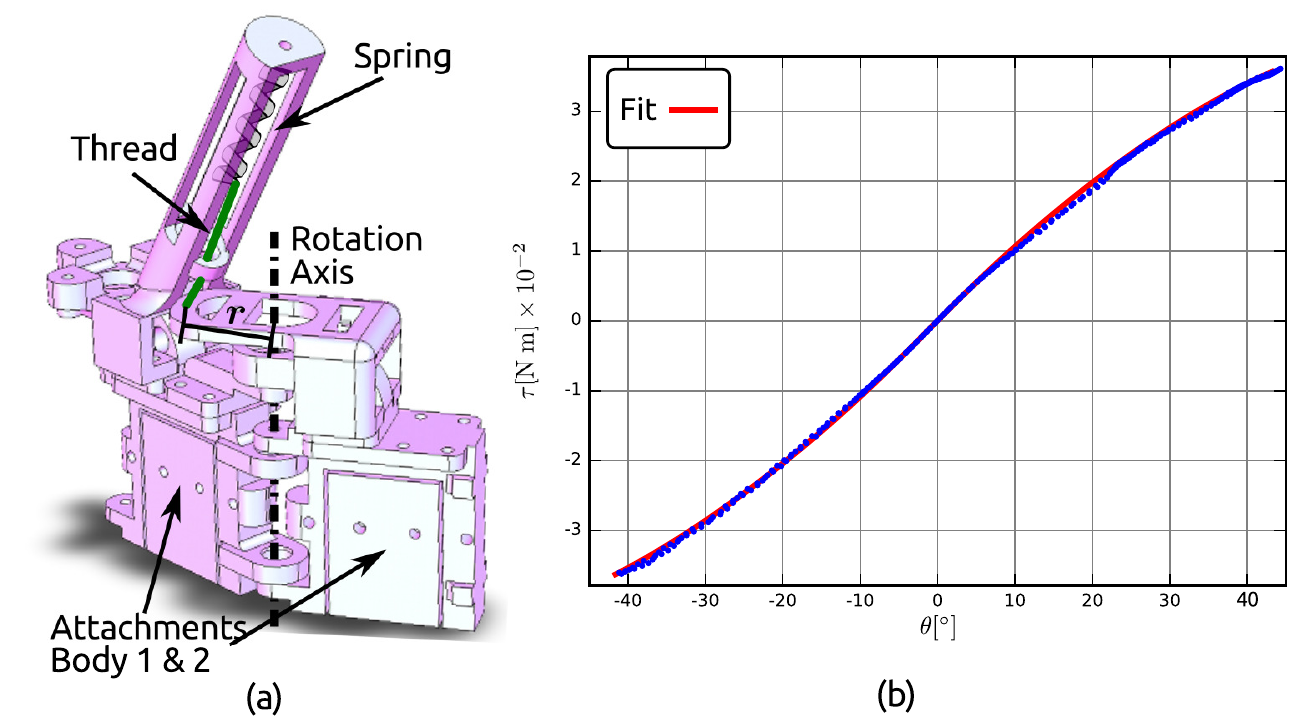}
\caption{Description of the joints in WandaX. {\bf a)} Details of the joints used in~\cite{Ziegler11}, note the axis of rotation and the appendage. {\bf b)} Measured torques applied to the joint under controlled deflection. Replacing the parameters given in Table~\ref{tab:Params} into Eq.~\eqref{eq:Torque} the fit provides $F = \unit{0.73 \pm 0.05}\newton$.}
\label{fig:JointCAD}
\end{figure*}

To extract energy from the environment, the robotic platform has to be optimally driven by interaction forces. In this case, by the interaction between the rigid body of the robot and the surrounding turbulent flow. The feat can not be performed if the angle trajectories of the joints are fully prescribed by the controller. Therefore, the joints of our robot are compliant, i.e. the motion of the joint is not only defined by the controller, but also by external actions. In Figure~\ref{fig:JointCAD}a we see the details of the joints of the robot fish used in~\cite{Ziegler11}. Each joint behaves as a rotational spring. The restoring torque is generated when the relative angle between the two connected bodies is not zero. The force producing the torque is given by the extension of a linear spring fixed to the first body. The spring is connected via an inelastic thread to an appendage of the second body. When the deflection angle is zero, the extension of the spring is minimum as well as the force it exerts. We call {\it tension} to this minimum force value  and it is referred with the letter $F$. Measurements of the torque for $F = \unit{0.73}\newton$, are given in Figure~\ref{fig:JointCAD}b. The values of the parameters used throughout this paper are given in Table~\ref{tab:Params}. The two parameters $r,d$ are distances that can be seen in the figures. The elastic constant of the linear spring is $K$ and $I$ denotes the moment of inertia of the joint around the axis of rotation. The linear specific damping coefficient of the joint is denoted with $\zeta$. The parameters $\mathcal{Q}_0$ and $\Omega$ correspond to the amplitude and frequency of the external forcing, respectively.

\begin{table}[htb]
\centering
	\begin{tabular}{|ll|}
	    \hline
		Name & Value \\ \hline
		$r$ & \unit{20.24\pm 0.02}\milli\metre \\
		$d$ & \unit{27.68\pm 0.02}\milli\metre \\
		$K$ & \unit{81\pm 1}\newton\per\metre \\
		$I$ & \unit{(3.1 \pm 0.1)\times 10^{-5}}\kilogram\usk\metre\squared\\
		$\zeta \cdot I$ & \unit{(2.2 \pm 0.1)\times 10^{-4}}\newton\usk\metre\usk\second \\
		$\mathcal{Q}_0 \cdot I$ & \unit{1 \times 10^{-4}} \newton\usk\metre \\
		$\nicefrac{\Omega}{2\pi}$ & \unit{\left(0,3\right]}\hertz \\
		 \hline
	\end{tabular}
	\caption{Value of the parameters used here and in the model studied in~\cite{Ziegler11} .}
	\label{tab:Params}
\end{table}
%\end{flushleft}

\section{\label{sec:model}Mathematical model}
\begin{figure}[tb]
\centering
\includegraphics[width=.45\textwidth]{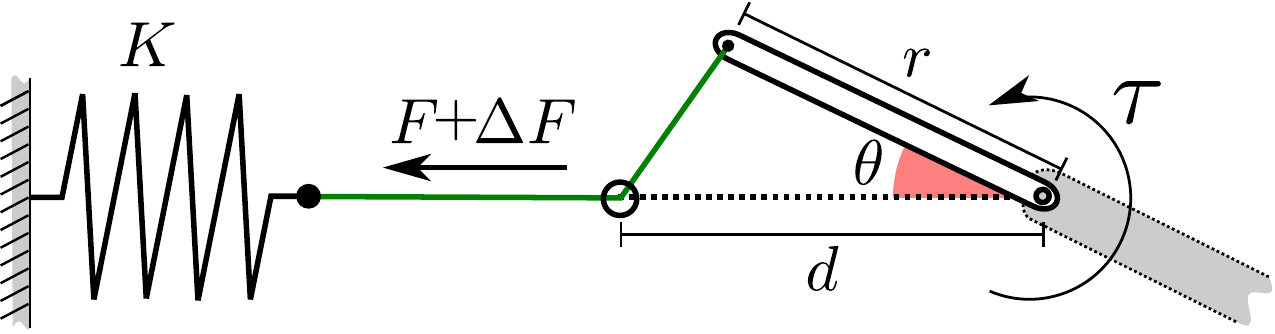}
\caption{Schematic of the rotational spring used to derive Eq. \eqref{eq:Torque}.}
\label{fig:JointModel}
\end{figure}

A geometrical representation of the joint is given in Figure~\ref{fig:JointModel}. The parameters $r$,$d$ and $K$ are fixed at construction time and always $d>r$ (Table~\ref{tab:Params}). The tension in the spring at its shortest length ($F$), is the controlled parameter and a servomotor can change it dynamically.

The torque applied to the bodies connected to the joint is thus,
\begin{equation}
\begin{split}
\tau =& \\
&\frac{K \left(\sqrt{\epsilon^2 + 4S\sin^2 \frac{\theta}{2}} - \epsilon \right) + F}{\sqrt{\epsilon^2 + 4S\sin^2 \frac{\theta}{2}}} S\sin\theta .
\end{split}
\label{eq:Torque}
\end{equation}
\noindent Where we have defined the parameters $\epsilon = d - r$ and $S = rd$. The formula is obtained by calculating the deformation of the spring as a function of the deflection angle of the appendage ($\theta$). The deformation is inside parenthesis in the numerator of equation~\eqref{eq:Torque}  and when multiplied by the stiffness $K$, it gives the force due to the deformation of the linear spring. The outer factors come from the product of the force and the moment arm. Note that the the torque $\tau$ is linear in the controlled input $F$.

For $\theta \ll 1$ the third order Taylor expansion gives
\begin{equation}
\tau(\theta,F) = \kappa(F)\theta + \alpha(F)\theta^3 + \mathcal{O}(\theta^5).
\label{eq:Taylor}
\end{equation}
\noindent Where
\begin{eqnarray}
\kappa(F) &=& \frac{SF}{\epsilon}, \\
\alpha(F) &=& \frac{SF}{\epsilon}\left[\frac{S}{2\epsilon^2}\left(\frac{\epsilon K}{F} - 1\right) - \frac{1}{6}\right].
\end{eqnarray}
\noindent And the equation of motion of the deflection angle is
\begin{equation}
\begin{split}
\ddot{\theta} + \zeta \dot{\theta} + \frac{\tau(\theta,F)}{I} &\approx \\
\ddot{\theta} + \zeta \dot{\theta} + k(F)\theta &+ a(F)\theta^3 \; = \;\Gamma.
\end{split}
\label{eq:DuffingO}
\end{equation}
\noindent Where $I$ denotes the moment of inertia around the axis of rotation. The specific damping is given by $\zeta$, and we have defined $k = \nicefrac{\kappa}{I}$, $a = \nicefrac{\alpha}{I}$. $\Gamma$ is the specific net effect of all other external torques acting on the joint. The approximating equation of motion is the well studied D\"uffing's equation.

\begin{figure}[htb]
\centering
\includegraphics[width=.45\textwidth]{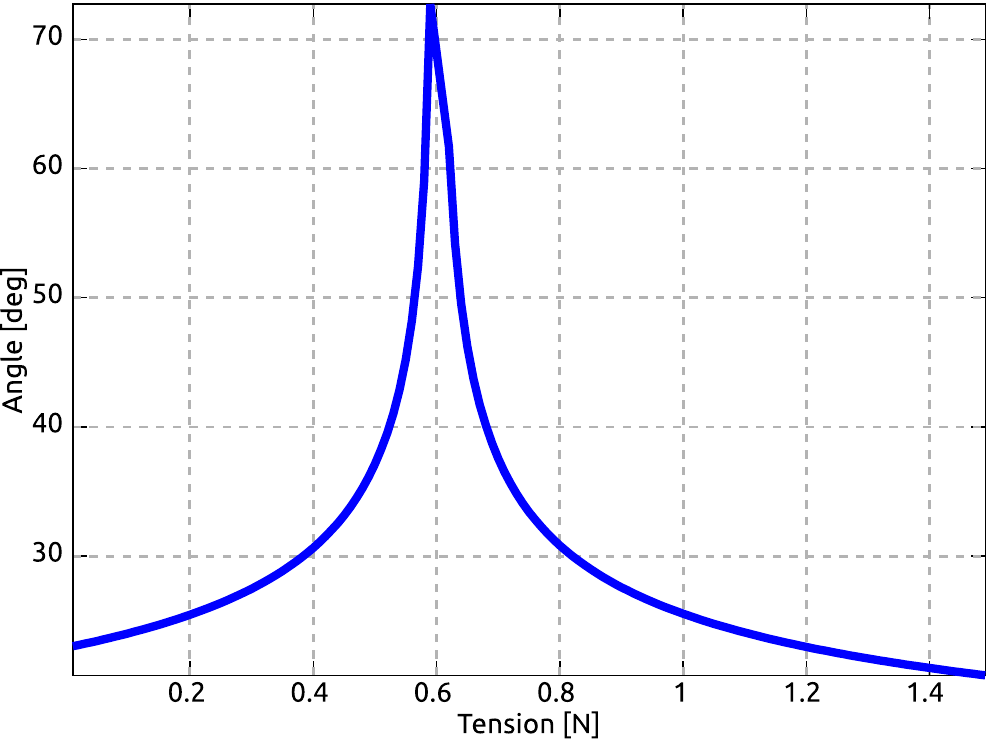}
\caption{Approximation error. Values of the deflection of the joint for which the error reaches the reference value given in the text. The maximum deflection has a peak close to $F_3 = \unit{(5.90\pm 0.01) \times 10^{-1}}\newton$ meaning that terms of degree $> 3$ almost cancel each other.}
\label{fig:ErrorModel} \end{figure}

To quantify the error introduced by the approximation, we calculated the angle at which the difference between the torque produced by equation \eqref{eq:Torque} and equation \eqref{eq:Taylor} is equal to a reference error given by $\Delta\tau = r\Delta F$, where $\Delta F = \unit{0.05}\newton$ is a reasonable resolution for a force sensor working in a $\unit{10}\newton$ range. These angles are plotted in Figure~\ref{fig:ErrorModel} for different values of the tension.

\subsection{Hardening, linear and softening spring}
\begin{figure}[htb]
\centering
\includegraphics[width=.45\textwidth]{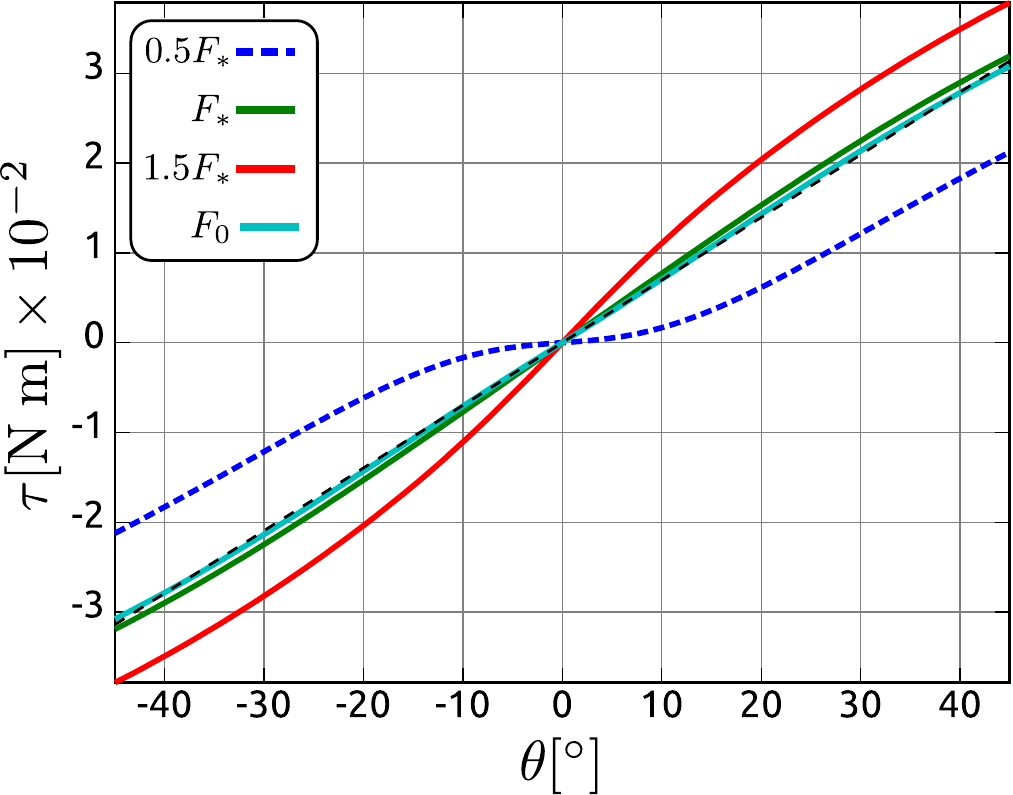}
\caption{Plots of the torque function for values of tension $F =[0.5,\,1,\,0.9,\,1.5]\cdot F_*$. Note the hardening behavior for small tension and softening behavior for higher tension. Almost linear curves are found for $F=F_*$ and $F=F_0$, as a reference the linear curve is shown.}
\label{fig:JointTorque} \end{figure}

As can be seen from equation \eqref{eq:Taylor}, $\alpha(F)$ modulates the intensity of the cubic non-linear term. This term vanishes when

\begin{equation}
F = F_* = \frac{\epsilon K}{\cfrac{\epsilon^2}{3S}+1}.
\end{equation}
\noindent rendering a linear spring for the angles where the third order approximation is valid. For bigger values of $F$ the spring will be softening ($\alpha<0$) and for smaller values it will be a hardening spring ($\alpha>0$) (Fig.~\ref{fig:JointTorque}). However, the full expression of the torque (Eq. \eqref{eq:Torque}) contains higher order terms. Hence, the linear behavior will be even more evident if the higher order terms cancel each other. In Figure~\ref{fig:JointTorque} curves of torque versus angle for several values of $F$ are shown. In particular we show the curve for which up to seventh order nonlinearities give a minimum contribution (found via optimization $F_0 = 0.9F_*$) together with the curve at $F = F_*$. These illustrates the power of the actuation chose, since we can control the dynamical properties of a virtual rotational spring at the joint (more details in~\cite{Ziegler11}).

\subsection{Forcing model}
As explained before, the external torques acting on the joint are due to fluid-structure interaction. This kind of interaction still poses a great challenge for the mathematical modeling community. In the search for simplified models of the forces generated in the interaction between flexible bodies and turbulent flows, we found the work of~\cite{Kanso09} and~\cite{Alben09} instructive. Using equation (3.8) given by~\cite{Alben09}, we can calculate the pressure difference on the boundary of a slender body in a vortex street. If the width of the body is bigger than the separation among vortices and it is placed in the middle of the vortex street, the forces on the body can be approximated by a sine function $f(t) = f_0 \sin(\Omega t + \phi)$, where $t$ is time and $\phi$ is an initial offset that sets the balance between the sine and cosine behavior of the forcing. The frequency $\Omega$ is proportional to the speed of the flow plus the vorticity of the vortices (assumed to be equal for each vortex) scaled by a factor that depends on the geometrical properties of the wake. The amplitude $f_0$ is proportional to the density of the fluid and the square of the vorticity of each vortex. With a few additional assumptions, the torque acting on the joint can be made proportional to this force. Here we adopt this over-simplified forcing model to avoid diverting the attention of the reader from the core ideas of our work. In this manner we postpone a detailed study with a more elaborated forcing model.

\section{\label{sec:SolMeth} Solution methods}
In this section we present two independent methods to estimate the amplitude of oscillations of the joint under periodic forcing. The first method turns out to be excellent for estimating the amplitude of the first harmonic. The second method is based on a Volterra series expansion of the Duffing equation. It allows to study the response of the joint to more general forcing conditions, but the tension range for which it is valid is smaller.
\subsection{\label{sec:hsDuff}Harmonic solutions of D\"uffing's equation}
Under periodic forcing $\Gamma = \mathcal{Q}_0 \sin(\Omega t)$, equation \eqref{eq:DuffingO} has been extensively studied (see ~\cite{Holmes76,Luo1997} and references therein). Following these analyses, we show here how to maximize the amplitude of the periodic response of the joint by tunning the tension parameter.

The key point of the analysis in~\cite{Luo1997}, is that we search for the amplitude of solutions of \eqref{eq:DuffingO} that are periodic (this rules out sub-harmonics, supra-harmonics and chaotic motion), $\theta(t) = A \sin(\Omega t + \psi)$. Under this assumption it can be shown that the amplitude $A$ of such solutions is given by the roots of the polynomial,
\begin{equation}
\begin{split}
A^3 \frac{9a^2}{16} + A^2 \frac{3\left(k - \Omega^2\right)a}{2} +& \\
A\left[\left(\Omega^2 + \zeta^2-2k\right)\Omega^2 + k^2\right] &- \mathcal{Q}_0^2 = 0.
\end{split}
\label{eq:Amplitude}
\end{equation}
\noindent The roots of this polynomial can be obtained analytically using a computer algebra system as~\cite{Maxima} and they establish the relation between the amplitude of the oscillations and the parameters of the equation, $A\left(k,a,\Omega,\mathcal{Q}_0\right)$. In the case at hand, we have $k(F)$ and $a(F)$, therefore $A\left(F,\Omega,\mathcal{Q}_0\right)$. For a given value of the parameters, only one root corresponds to the observed amplitude (there are unstable amplitudes). This implies, that is not enough to look at the roots, but we must also check their stability. This adds some complexity to the evaluation of the results that adds to the limitation of pure harmonic inputs.

\subsection{\label{sec:volterra} Volterra series expansion}
When the amplitude of the forcing is sufficiently small, the behavior of Duffing's oscillator in the neighborhood of the origin can be described by polynomial Volterra functionals(Theorem 3.1 of~\cite{Rugh1981}). This means that the angle of the oscillations can be described by an expansion of the form
\begin{equation}
\theta(t) \approx \sum_{i=1}^n y_i(t),\label{eq:Voltexpan}
\end{equation}
\noindent where $n$ is the order of the expansion and each term is the multi-dimensional integral
\begin{multline}
y_i(t)  = \int_{-\infty}^{+\infty}h_i(t-\sigma_1,\ldots,t-\sigma_i)\\
\Gamma(\sigma_1)\ldots \Gamma(\sigma_i)\ud\sigma_1\ldots\ud\sigma_i.\label{eq:outputIntegral}
\end{multline}
\noindent where $\Gamma(t)$ is the forcing signal and we wish to determine the kernel functions $\left\{h_i\right\}_{i=1}^n$. However, for the objective at hand, it is more useful to calculate the Fourier transform $Y(\omega)$ of these functionals. We have not found this process described in the literature, hence we describe it succinctly. We used the inverse Fourier transform definition on the input signal and substitute it into the functional definition Eq.~\eqref{eq:outputIntegral}. By inverting the order of integration and splitting variables, we make the kernel transform $H(\omega)$ appear. Then, we compare this expression to the formal definition of the inverse Fourier transform of $Y(\omega)$ and isolate the desired result. This process can be applied for the first three orders, but it gets cumbersome for higher ones. Therefore, before showing the results, we will briefly introduce some notation used here to simplify the presentation.

An ordered set of arguments $(s_1,s_2,\ldots,s_n)$ will be denoted $s_{1:n}$. In general we have,
\begin{equation*}
(s_k,s_{k+1},\ldots,s_n) = s_{k:n} \quad k \leq n.
\end{equation*}
\noindent For example, the fifth order kernel $H_5(s_1,s_2,s_3,s_4,s_5)$ will be written
\begin{align*}
&H_5(s_{1:5})  = \\
& -3a\, H_1\left(\Sigma s_{1:5}\right)H_1(s_1)H_1(s_2)H_3(s_{3:5}),
\end{align*}
\noindent where $\Sigma s_{1:5} = \sum_{i=1}^5 s_i$. For the integration variable in multiple integrals we will write $\ud s_{1:n} = \ud s_1\ud s_2 \cdots \ud s_n$. For multiple summations over $n$ indexes we will write $\sum_{k_{1:n}} = \sum_{k_1}\ldots \sum_{k_n}$. Having defined the notation, we continue with our presentation.

For zero initial conditions, $(\theta,\dot{\theta})=0$, and working in the frequency domain, each kernel can be obtained recursively based on lower order kernels (details of the calculation are given in~\cite{PeytonJones}).
The first order kernel derived from the model in Eq.~\eqref{eq:DuffingO}, is equivalent to the frequency response function (also known as transfer function) of a linear second order system,
\begin{equation}
H_1(s)  =  \frac{1}{s^2+\zeta s+k}\label{eq:H1}.
\end{equation}
\noindent The following equations present the Volterra kernels up to seventh order using recurrent relations,
\begin{gather}
\begin{gathered}
\begin{split}
H_3(s_{1:3})  =  -a &H_1\left(\Sigma s_{1:3}\right)\cdot\\
&H_1(s_1)H_1(s_2)H_1(s_3),
\end{split}\\
\begin{split}
H_5(s_{1:5})  =  -3a &H_1\left(\Sigma s_{1:5}\right)\cdot\\
&H_1(s_1)H_1(s_2)H_3(s_{3:5}),
\end{split}\\
\begin{split}
H_7(s_{1:7})  = &-3a H_1\left(\Sigma s_{1:7}\right)\cdot\\
& \Big[\; H_1(s_1)H_1(s_2)H_5(s_{3:7})+ \\
&  H_1(s_1)H_3(s_{2:4})H_3(s_{5:7})\; \Big].
\end{split}
\end{gathered}\label{eq:RecursiveRel}
\end{gather}
\noindent Due to the symmetry of the eq.~\eqref{eq:DuffingO}, all even order kernels are zero. Note that these kernels are valid for any second order cubic oscillator. The recursive formulas were obtained using the probing method of~\cite{PeytonJones}, which proceeds as follows. We start from a polynomial nonlinear ordinary differential equation (as Eq.~\eqref{eq:DuffingO}) and assume the response can be represented by Volterra series, i.e. Eq.~\eqref{eq:Voltexpan}. We substitute this into the system's differential equation, and equate similar terms. This last step produces equations relating different order kernels and the expressions are always recursive (we obtain Eqs.~\eqref{eq:RecursiveRel} in our case). The factorial in the recursive relations appear from the multinomial expansion of nonlinear polynomial terms.

We proceed to calculate the Fourier transform of the output given by Eq.~\eqref{eq:outputIntegral} subject to harmonic forcing. We will use Eqs.~\eqref{eq:RecursiveRel} in order to obtain an explicit relation between the parameters of our model $k,a,\zeta,\mathcal{Q}_0,\Omega,F$ and the amplitude of the first harmonic of the output.

\begin{figure*}[t!]
\centering
\subfloat[]{\label{fig:JointORFLuo}\includegraphics[width=.45\textwidth]{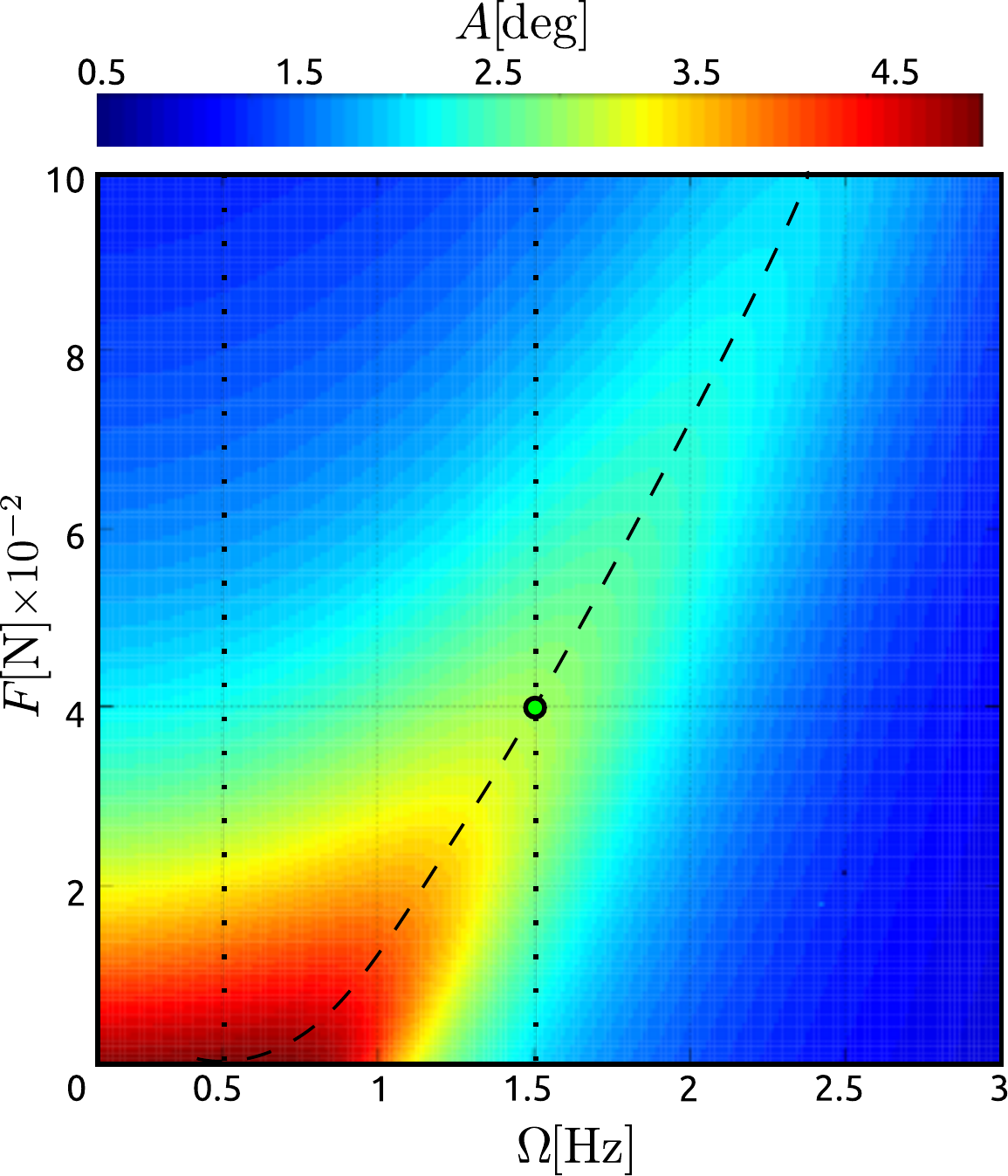}}
\subfloat[]{\label{fig:ORFError}\includegraphics[width=.43\textwidth]{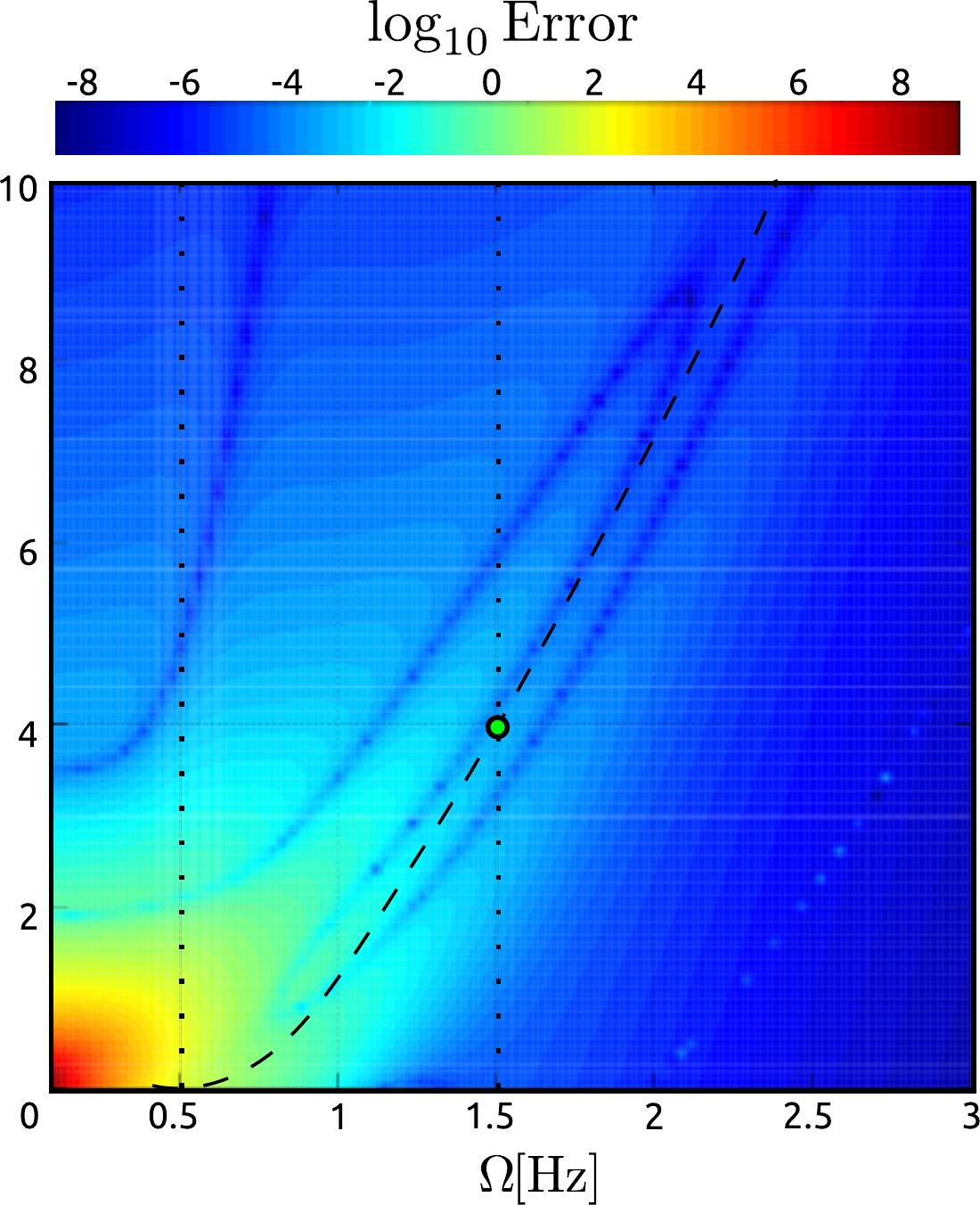}}
\caption{\label{fig:ORF}Amplitude of periodic response in the plane $(F,\Omega)$. \protect\subref{fig:JointORFLuo} Amplitude of the oscillations according to Eq.~\eqref{eq:Amplitude}. The line of maxima is marked with dashes. The vertical dotted lines indicate selected values of $\Omega$. \protect\subref{fig:ORFError} Relative differences between the predicted amplitudes by the method described in section~\ref{sec:hsDuff} and the Volterra expansion described in section~\ref{sec:volterra}. }
\end{figure*}

We recall that the Fourier transform of a general bandwith limited periodic function is a scaled Dirac Comb (Shah function, Impulse train, Dirac train, etc.) (see ~\cite{Schetzen1980}),
\begin{align}
\gamma\left(\omega\right)  =  \sum_{l=-L}^{+L}X_{i}\delta\left(\omega-\Omega l \right)\label{eq:DFT}
\end{align}
\noindent where $L$ is a positive integer, $X_i$ is the complex amplitude of the $i$-th harmonic and $\delta(\cdot)$ is the impulse function (Dirac delta function).
The Fourier transform of each of the output terms in~\eqref{eq:Voltexpan} is
\begin{equation}
\begin{gathered}
Y_{i}(\omega) = \left(2\pi\right)^{1-i}\int_{-\infty}^\infty H_i\left(\Delta\omega_i\right) \cdot \\
\gamma(w_1)\cdots\gamma(w_{i-1})\gamma\left(\omega-\Sigma w_{1:i-1}\right)\ud w_{1:i-1} =\\
(2\pi)^{1-i}\sum_{l_{1:i}=-L}^{+L}
B_{l_1:l_i}\delta\left(\omega-\Omega\Sigma l_{1:i}\right),
\end{gathered}\label{eq:FToutput}
\end{equation}

\noindent where we have used Eq.~\eqref{eq:DFT} and $\Delta\omega_i = (w_1,\ldots,w_{i-1},\omega-\Sigma w_{1:i-1})$ and,
\begin{align}
B_{l_1:l_i} = H_i\left(\Omega l_{1:i}\right)X_{l_1}\cdots X_{l_i}.
\end{align}
The integral in~\eqref{eq:FToutput}, has an interesting geometric interpretation in terms of convolutions in hyperplanes, we refer the interested reader to~\cite{Lang1996}. Therein, the output frequency range of nonlinear systems that are representable by Volterra series is analytically calculated. Additionally, in Eq.(16) of that paper the frequency spectrum of the output signal is represented as the superposition of contributions from the nonlinearities.
In the case studied herein, the input has only one frequency, therefore $L=1$, $X_0=0$, $X_{\pm 1}=\pm j\mathcal{Q}_0\pi$ with $j$ the imaginary unit. Replacing these values in all the equations and using the relations in Eq.~\eqref{eq:RecursiveRel} we obtain the desired result,
\begin{multline}
Y(\Omega)= \frac{-j\pi}{64} \Big\{
3a^3\mathcal{Q}_0^7 \big[ \phantom{aaaaaaaaaaaaaaaaaaaaaaaaa}\\
2 H_1(3\Omega)H_1(-3\Omega)H_1^3(-\Omega)H_1^5(\Omega) \\
+ 6  H_1(3\Omega)H_1^4(-\Omega)H_1^5(\Omega) \phantom{aaaaaaa}\\
+ 6  H_1^2(3\Omega)H_1^3(-\Omega)H_1^5(\Omega)\phantom{aaaaaaa}\\
+ 3  H_1(-3\Omega)H_1^4(-\Omega)H_1^5(\Omega)\phantom{aaaaaa}\\
- 15  H_1(3\Omega)H_1^3(-\Omega)H_1^6(\Omega)\phantom{aaaaaaa}\\
+ 45 H_1^3(-\Omega)H_1^7(\Omega)\phantom{aaaaaaaaaaaa}\\
+ 45 H_1^4(-\Omega)H_1^6(\Omega)
+ 18 H_1^5(-\Omega)H_1^5(\Omega) \; \big] \phantom{aaaaaaaaaaa}\\
- 12a^2\mathcal{Q}_0^5 \big[
H_1(3\Omega)H_1^2(-\Omega)H_1^4(\Omega) \phantom{aa}\\
+ 6 H_1^2(-\Omega)H_1^5(\Omega) \phantom{aaaaaa}\\
+ 3 H_1^3(-\Omega)H_1^4(\Omega) \; \big] \phantom{aaaaa} \\
+ 48a\mathcal{Q}_0^3 H_1(-\Omega)H_1^3(\Omega) - 64\mathcal{Q}_0H_1(\Omega) \Big\}.\label{eq:OutputFT}
\end{multline}
\noindent Where $H_1(x)$ is given by Eq.~\eqref{eq:H1} with $s = jx$. The amplitude of the oscillation is obtanied by taking the double of the modulus of the complex number $Y(\Omega)/\pi$.

\section{\label{sec:orf} Results}

\begin{figure}[htb]
\includegraphics[width=.48\textwidth]{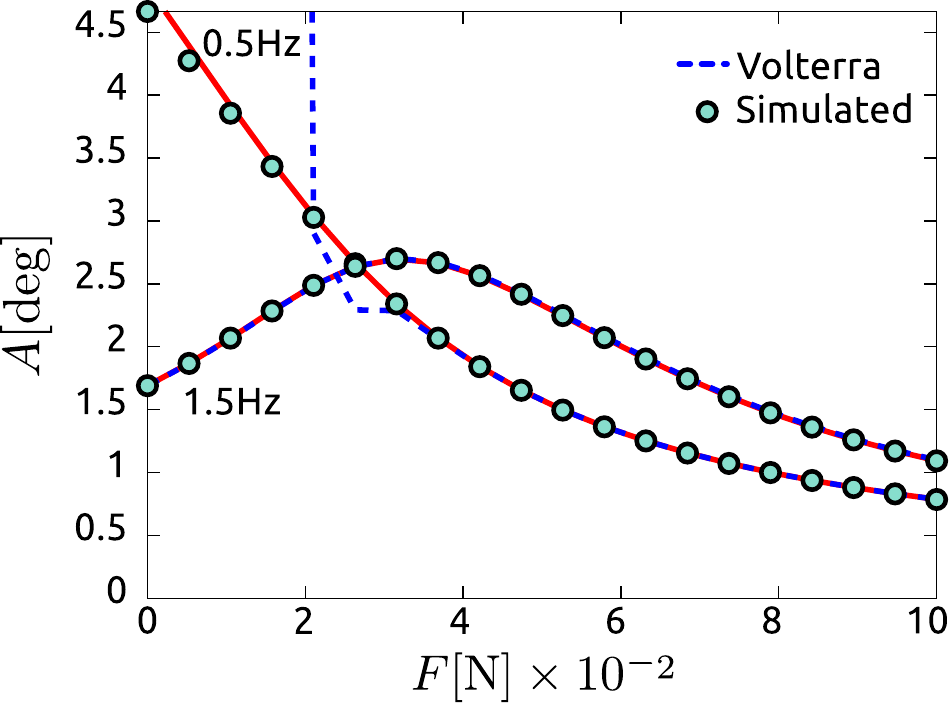}
\caption{\label{fig:ORFcut} Amplitude of periodic response of the joint for forcing with frequencies $\Omega = [\unit{0.5, 1.5}]\hertz$. Amplitudes according to Eq.~\eqref{eq:Amplitude} in solid line, amplitudes obtained from the Volterra expansion Eq.~\eqref{eq:OutputFT} in dashes and amplitudes from simulations without approximation in circles.}
\end{figure}

Figure~\ref{fig:JointORFLuo} shows the amplitude of periodic oscillations in the $(F,\Omega)$ plane, according to Eq.~\eqref{eq:Amplitude}. The line of maxima is shown with dashes. This line describes the value of the tension that produces maximum amplitude for a given forcing frequency. In Fig.~\ref{fig:ORFError} we plot the logarithm of the relative difference between the amplitudes given by~\eqref{eq:Amplitude} and~\eqref{eq:OutputFT}. The two approximation differ for regions of low frequency and low tension where the system is most nonlinear~(\cite{Luo1997}). To compare with the observed amplitude (obtained by simulating~\eqref{eq:DuffingO} without any approximation), we extracted the curves of amplitude against tension for $\nicefrac{\Omega}{2\pi} = \unit{0.5, 1.5}\hertz$ (vertical dotted lines in Fig.~\ref{fig:ORF}). These curves are shown in Figure~\ref{fig:ORFcut} together with the amplitude calculated using Eq.~\eqref{eq:OutputFT} corresponding to the Volterra kernel expansion. For the $\unit{1.5}\hertz$ frequency, both models predict the simulated amplitude accurately. For the $\unit{0.5}\hertz$ frequency, the amplitude predicted by Eq.~\eqref{eq:Amplitude} drifts away from the simulated value for lower tensions. The amplitude calculated using the Volterra expansion diverges for low tensions at this frequency.
In the same figure we show the amplitudes measured in simulations using the exact expression for the torque~\eqref{eq:Torque} (Octave~(\cite{octave2002}) function {\tt ode45}, absolute and relative tolerances, $10^{-6}$). The amplitude of the oscillations is obtained using the absolute value of the analytic signal of the angle (Octave function {\tt hilbert}). The agreement between the approximated results and the simulated ones is noteworthy.

These results show that the approximation from~\cite{Luo1997} can be used for a tension controller designed for a joint of this kind, to obtain periodic responses with the same frequency as the forcing. The Volterra approximation fails when the system becomes strongly nonlinear, however these expansion can be used for more general responses or when the inputs have a more complicated frequency spectrum (as in a realistic scenario).

\section{\label{sec:conclu}Discussions and conclusion}
Though the quote from~\cite{Brilliant58} remains valid, we used the knowledge about D\"uffing's equation to understand analytically a compliant joint. This gives a corner stone to study more complicated setups (e.g. chain of joints, as in the robotic fish) and define the robust engineering design of robots with the desired resonance properties. Additionally, the results showed here provide a reference solution to the problem of finding the right pretension for a given external forcing, related to the idea of adaptive controllers. For a forcing with a sufficiently slow varying frequency, the tension could be adjusted to maintain the amount of energy transferred into the system to its maximum possible value, i.e. keep the system close to the line of maxima in Fig.~\ref{fig:JointORFLuo}. With the results herein the adjustment could be done by direct calculations. However, other methods like the frequency oscillators(AFO)~(\cite{Buchli06}) were used to achieve a similar objective. Applications of the latter has been reported in~\cite{Buchli08}, however, due to lack of theoretical results on the resonance frequency of the nonlinear platform studied therein, the theoretical solution to the problem was unknown and a grounded evaluation of the performance was not possible. Approximations as the one presented here could be used to verify the results generated by the adaptive oscillators.

In our results we exclude any information about oscillations with frequencies different from that of the forcing since that is the case where Eq.~\eqref{eq:Amplitude} is valid. The higher order harmonics in the system's output could be determined and compared with the simulated results. Traditional harmonic balance method can not be used to find these results. Additionally, a more complete panorama of the frequency response of the system could be obtained with a nonlinear analysis method as the NOFRF (proposed in~\cite{Lang2005} and showcased in~\cite{Peng2007}), which is based on Volterra series expansions as the one presented in this work. NOFRF give the response of the system to a given set of inputs, therefore the {\it natural ensemble} of forcing signals must be known to acquire useful information. This ensemble has to be compiled from (or modeled based on) fluid-structure interactions data. Comparing the present results with the information provided by NOFRF will be the objective of our next work. Proved that such analysis is helpful for the design of compliant robots, we will extend the analysis to biped and quadruped robots. In that situation the forcing comes from the ground reaction forces, that are generally not smooth (impacts) and contain a rich frequency spectrum. We consider those scenarios as highly complex problems to be addressed with a mature toolbox of methods.

Concluding, we have provided a method that allows the construction of a controller that would maximize the harvesting of environmental energy, under the circumstances defined herein. The method will be extended to cover more realistic situations than the simplified periodic forcing, including forcing with wider frequency spectrum. Additionally, the same methodology presented here can be used to verify results obtained using heuristic methods as AFO or neural networks. Our results can be easily validated and we invite our colleagues to do it.

\section*{Acknowledgements}
The authors would like to thank Cecilia Tapia Siles for sharing insights and information about passive swimmers.
\paragraph*{Funding.} The research leading to these results has received funding from the European Community's Seventh Framework Programme FP7/2007-2013-Challenge 2 --Cognitive Systems, Interaction, Robotics-- under [grant agreement No 248311-AMARSi]. RB is founded by the CAPES Foundation, Ministry of Education of Brazil. MZ received funding for this work from the SNSF [project no. 122279] (From locomotion to cognition).
\paragraph*{Author contributions.} JPC, RB and ZQL worked on the mathematics of the models and in the simulations. MZ designed and built the joint studied in this work. All authors collaborated for the production of the manuscript.

\bibliographystyle{cell}
\bibliography{../references}

\end{document}